\pdfoutput=1
\documentclass{article}
\usepackage{ifpdf} 
\usepackage{spconf,amsmath,hyperref}
\usepackage[pdftex]{graphicx}
\usepackage{colortbl}
\usepackage{xcolor}
\usepackage{multirow}

\usepackage{booktabs}
\usepackage{url}
\usepackage{epstopdf}
\usepackage{amssymb}

\title{PPOM: MARGINALIZING PATCH-GRID PHASE FOR CLIP-BASED GENERALIZABLE VISION-LANGUAGE PROMPT TUNING}
\name{Liang Wang$^{1,2}$, Haoyang Li$^{1,2}$, Chao Wang$^{1,*}$, Guodong Long$^{2,*}$, Jing Jiang$^{2,*}$, Yan Peng$^{1,}$\sthanks{$^{}$Corresponding authors.}}
\address{$^{1}$Shanghai University, Shanghai, China \quad $^{2}$University of Technology Sydney, Sydney, Australia}
\begin{document}
%
\maketitle
\begin{abstract}
Prompt tuning adapts CLIP-based vision-language models with few trainable parameters, yet its predictions remain sensitive to the spatial sampling imposed by a frozen vision transformer. In particular, non-overlapping patch tokenization makes predictions depend on the alignment (\textbf{phase}) between image and the patch lattice. To reduce prediction sensitivity to patch-grid alignment, we introduce \textbf{P}atch-\textbf{P}hase \textbf{O}rbit \textbf{M}arginalization (\texttt{PPOM}), a training-free inference operator that treats phase shift as a nuisance variable. Given a patch stride, PPOM evaluates the identity view and reflection-padded translations, pairs opposite shifts into horizontal, vertical, and diagonal antithetic families, and assigns equal mass to these families and the identity prediction to avoid view-count bias during phase integration. In summary, PPOM provides a deterministic interface between prompt adaptation and patch-grid sensitivity. Across multiple prompt-learning hosts, PPOM improves host performance without re-training.
\end{abstract}


\section{Introduction}



CLIP transfers visual recognition through the similarity between image and text representations \cite{Radford2021}. Building on this paradigm, prompt learning replaces fixed prompt templates with lightweight learnable tokens while keeping the CLIP backbone frozen \cite{Zhou2022CoOp}, enabling efficient adaptation to target (base) tasks while preserving generalization to unseen (new) tasks. Existing studies improve prompt learning from multiple perspectives, including prompt design \cite{Li2025DPC, wu2024caspl}, cross-modal prompting \cite{Khattak2023MaPLe, li2025augpt}, objective design \cite{Yao2023KgCoOp, Khattak2023PromptSRC, zhang2024dept}, mid-layer plugin modules \cite{Yang2024MMA, Guo2025MMRL}, and external information augmentation \cite{li2025atprompt, li2026fvgpt}. Despite these advances, existing methods pay limited attention to the visual sampling and encoding process.

\begin{figure}[t]
  \centering
  \includegraphics[width=\linewidth]{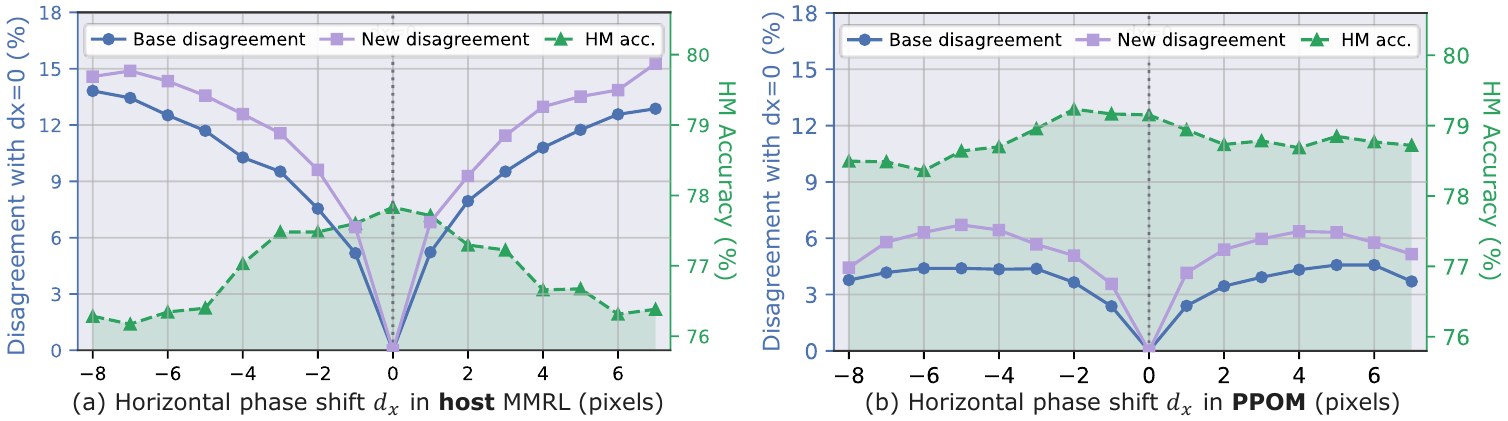}
  \caption{Impact of horizontal patch phase shift on the test set of prompt learner: (a) Host predictions show substantial phase disagreement rate, while (b) PPOM significantly reduces disagreement and improves harmonic mean (HM) accuracy.}
  \label{Fig-eyecatch}
\end{figure}

Specifically, the vision transformer (ViT) in frozen CLIP tokenizes an image into non-overlapping tokens on a fixed patch lattice \cite{Dosovitskiy2021}. A small translation can change the relative alignment between the image and this lattice (i.e., \textbf{phase}), and alter which pixels share a patch even when the semantic content is preserved \cite{RojasGomez2024}. As shown in Fig.~\ref{Fig-eyecatch}(a), shifting the image by only a few pixels substantially increases prediction disagreement and can reduce the accuracy of host prompt learners. This sensitivity means that \textbf{\textit{prompt learning optimizes semantic adaptation under only a fixed tokenization phase}}, leaving the classifier exposed to phase-dependent prediction variance at inference. Consequently, even a well-trained prompt cannot explicitly remove the sampling-origin dependency inherited from the frozen visual encoder.

To mitigate patch-grid-induced interference at test time, we introduce \textbf{Patch-Phase Orbit Marginalization (PPOM)}. Unlike generic test-time augmentation \cite{lyzhov2020test-time} based on manually chosen transforms, PPOM isolates the architecture-defined patch-phase variable induced by the frozen CLIP backbone, requiring neither policy selection nor an additional tuning stage. Its support discretely approximates the patch-grid phase cell with identity, horizontal, vertical, and diagonal families; the displacement is fixed by the ViT patch stride, while equal family mass avoids over-weighting families with more enumerated views. This design preserves the prompt learner's adaptation while reducing phase-dependent prediction variance at inference. Our contributions are:
\begin{itemize}
    \item We formulate the patch-grid phase as an inference nuisance inherited from frozen ViT tokenization, complementary to semantic prompt adaptation.
    \item We propose PPOM, a test-time operator with an architecture derived phase orbit and family-balanced marginalization to reduce patch-grid sensitivity.
    \item Evaluation across 4 hosts shows that PPOM improves host-level performance without re-training.
\end{itemize}



\begin{figure*}[t]
  \centering
  \includegraphics[width=0.9\textwidth]{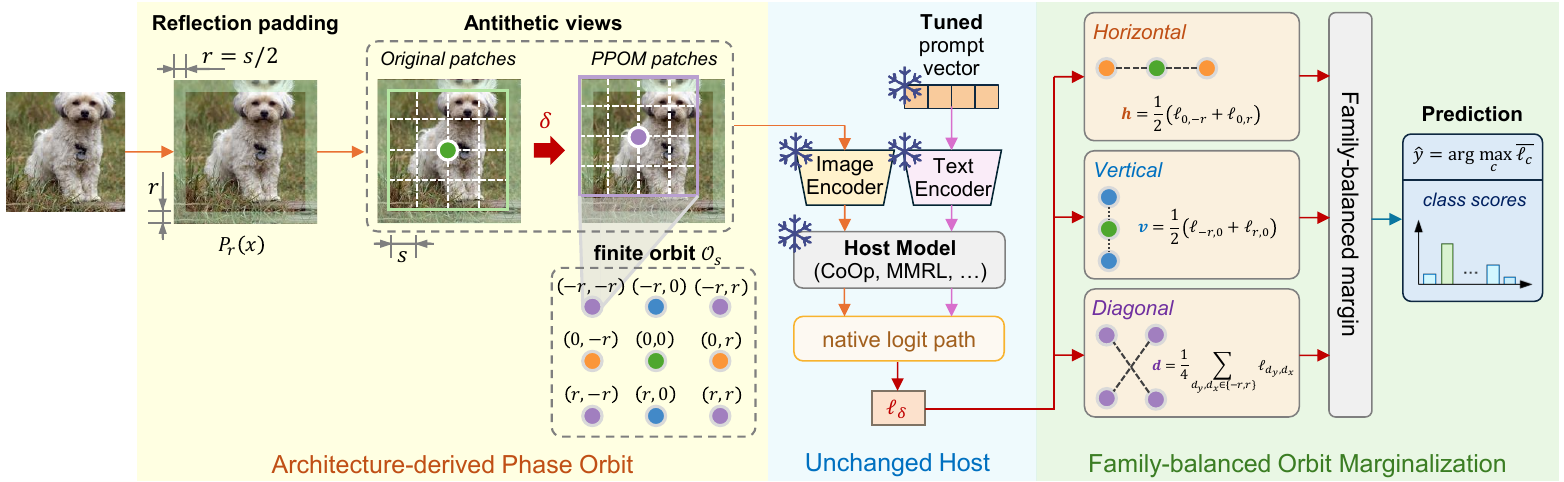}
  \caption{Framework of PPOM. \textit{Architecture-derived Phase Orbit} (Sec.~\ref{sec.2.2}) constructs stride-defined patch phases via reflection padding and half-patch translations $\delta$, evaluated by the unchanged prompt-tuned host. \textit{Family-balanced Orbit Marginalization} (Sec.~\ref{sec.2.3}) then assigns equal mass to identity, horizontal, vertical, and diagonal families for final prediction $\hat y$.}
  \label{Fig-framework}
\end{figure*}

\section{Patch-Phase Orbit Marginalization}

\subsection{Preliminaries}

Let frozen image and text encoders be $f_\theta$ and $g_\theta$. For class $c$, a generic prompt learner inserts trainable context $p$ around its class embedding $e_c$ to form $q_c(p)=[p_1,\ldots,p_M,e_c]$ \cite{Zhou2022CoOp}. Visual or intermediate prompts can be absorbed into the image path. The native logit of the host prompt learner is
\begin{equation}
\ell_H(x;p)_c=\tau^{-1}\operatorname{sim}\!\left(f_\theta(x;p),g_\theta(q_c(p))\right),
\end{equation}
The host learns $p$ only from a base-class set $\mathcal D_B$ by
\begin{equation}
p^*=\arg\min_p -\!\sum_{(x,y)\in\mathcal D_B}
\log\frac{\exp \ell_H(x;p)_y}{\sum_{c\in\mathcal C_B}\exp \ell_H(x;p)_c},
\end{equation}
while $\theta$ remains frozen. Visual prompts, intermediate tokens, and host-specific fused representations are absorbed into $\ell_H$. Plug-and-play PPOM is applied after the host has fixed $p^*$ and always calls the same native logit function.

\begin{table*}[t!]
    \caption{Base-to-new generalization performance (\%) of 4 host models w/ or w/o our PPOM on 11 datasets.}
    \label{tab:b2n}
\centering
\setlength\tabcolsep{2.5pt}
\small
\begin{tabular}{c|ccc|ccc|ccc|ccc|ccc|ccc}
\toprule
\rowcolor{gray!10} {\cellcolor{gray!10}} & \multicolumn{3}{c|}{\textbf{Average (11)}} & \multicolumn{3}{c|}{\textbf{ImageNet}} & \multicolumn{3}{c|}{\textbf{Caltech101}} & \multicolumn{3}{c|}{\textbf{Oxford Pets}} & \multicolumn{3}{c|}{\textbf{Stanford Cars}} & \multicolumn{3}{c}{\textbf{Flowers102}} \\
\rowcolor{gray!10} \multirow{-1.8}{*}{{\cellcolor{gray!10}}\textbf{Method}} & Base & New & HM & Base & New & HM & Base & New & HM & Base & New & HM & Base & New & HM & Base & New & HM \\
\midrule
CoOp & 80.91 & 70.01 & 75.07 & 75.96 & 68.93 & 72.27 & 97.81 & 94.65 & 96.20 & 95.11 & 97.04 & 96.06 & 68.97 & 69.67 & 69.32 & 96.77 & 71.42 & 82.18 \\
{\cellcolor{green!20}}\textbf{+PPOM} & {\cellcolor{green!20}}81.50 &
{\cellcolor{green!20}}70.59 &
{\cellcolor{green!20}}\textbf{75.65} & 76.93 & 69.49 & \textbf{73.02} & 98.32 & 94.76 & \textbf{96.51} & 95.48 & 97.26 & \textbf{96.36} & 70.61 & 70.46 & \textbf{70.54} & 96.58 & 72.13 & \textbf{82.58} \\
\midrule
KgCoOp & 80.32 & 72.16 & 76.02 & 76.18 & 70.26 & 73.10 & 98.06 & 95.63 & 96.83 & 95.06 & 97.93 & 96.47 & 71.84 & 74.77 & 73.28 & 95.92 & 71.42 & 81.87 \\
{\cellcolor{green!20}}\textbf{+PPOM} & {\cellcolor{green!20}}80.87 &
{\cellcolor{green!20}}72.54 &
{\cellcolor{green!20}}\textbf{76.48} & 77.08  & 70.78 & \textbf{73.80} & 98.39 & 96.07 & \textbf{97.21} & 95.37 & 98.15 & \textbf{96.74} & 73.44 & 75.61 & \textbf{74.51} & 96.49 & 72.13 & \textbf{82.55} \\
\midrule
PromptSRC & 81.25 & 73.66 & 77.27 & 77.00 & 70.60 & 73.66 & 97.93 & 94.54 & 96.21 & 95.32 & 97.43 & 96.36 & 71.51 & 75.51 & 73.46 & 96.87 & 73.83 & \textbf{83.79} \\
{\cellcolor{green!20}}\textbf{+PPOM} & {\cellcolor{green!20}}81.77 &
{\cellcolor{green!20}}74.24 &
{\cellcolor{green!20}}\textbf{77.82} & 77.96 & 71.10 & \textbf{74.37} & 98.32 & 94.98 & \textbf{96.62} & 95.48 & 97.43 & \textbf{96.44} & 72.61 & 76.80 & \textbf{74.65} & 96.87 & 73.76 & 83.75 \\
\midrule
MMRL & 85.51 & 75.33 & 80.10 & 78.52 & 70.84 & 74.48 & 98.52 & 94.43 & 96.43 & 95.32 & 97.32 & 96.31 & 82.11 & 73.98 & 77.83 & 98.67 & 76.95 & \textbf{86.47} \\
{\cellcolor{green!20}}\textbf{+PPOM} & {\cellcolor{green!20}}86.18 &
{\cellcolor{green!20}}75.91 &
{\cellcolor{green!20}}\textbf{80.72} & 79.41 & 71.42 & \textbf{75.20} & 98.64 & 94.32 & \textbf{96.44} & 95.69 & 97.76 & \textbf{96.72} & 83.31 & 75.39 & \textbf{79.15} & 98.58 & 76.95 & 86.43 \\
\midrule 
\midrule
\rowcolor{gray!10} {\cellcolor{gray!10}} & \multicolumn{3}{c|}{\textbf{Food101}} & \multicolumn{3}{c|}{\textbf{FGVC Aircraft}} & \multicolumn{3}{c|}{\textbf{SUN397}} & \multicolumn{3}{c|}{\textbf{DTD}} & \multicolumn{3}{c|}{\textbf{EuroSAT}} & \multicolumn{3}{c}{\textbf{UCF101}} \\
\rowcolor{gray!10} \multirow{-1.8}{*}{{\cellcolor{gray!10}}\textbf{Method}} & Base & New & HM & Base & New & HM & Base & New & HM & Base & New & HM & Base & New & HM & Base & New & HM \\
\midrule
CoOp & 90.14 & 91.46 & 90.80 & 35.47 & 30.95 & 33.06 & 80.26 & 73.84 & 76.92 & 78.13 & 48.67 & 59.98 & 88.45 & 45.87 & 60.41 & 82.94 & 77.61 & 80.18 \\
\textbf{+PPOM} & 90.67 & 91.71 & \textbf{91.19} & 37.52 & 30.29 & \textbf{33.52} & 80.43 & 74.20 & \textbf{77.19} & 78.13 & 48.91 & \textbf{60.16} & 88.45 & 48.77 & \textbf{62.87} & 83.35 & 78.47 & \textbf{80.84} \\
\midrule
KgCoOp & 90.67 & 91.87 & 91.26 & 34.39 & 35.15 & 34.77 & 80.63 & 77.30 & 78.93 & 78.70 & 56.64 & \textbf{65.88} & 77.76 & 47.82 & \textbf{59.22} & 84.33 & 75.01 & 79.40 \\
\textbf{+PPOM} & 91.24 & 92.28 & \textbf{91.75} & 36.55 & 36.29 & \textbf{36.42} & 80.95 & 77.61 & \textbf{79.25} & 77.89 & 56.76 & 65.67 & 77.76 & 46.64 & 58.31 & 84.38 & 75.66 & \textbf{79.79} \\
\midrule
PromptSRC & 90.70 & 91.71 & 91.20 & 35.17 & 34.31 & 34.74 & 81.72 & 77.46 & 79.53 & 80.32 & 56.52 & 66.35 & 83.60 & 59.82 & 69.74 & 83.61 & 78.58 & 81.02 \\
\textbf{+PPOM} & 91.14 & 92.21 & \textbf{91.67} & 37.52 & 35.21 & \textbf{36.33} & 81.95 & 77.80 & \textbf{79.82} & 80.09 & 57.49 & \textbf{66.93} & 83.59 & 60.18 & \textbf{69.98} & 83.92 & 79.66 & \textbf{81.74} \\
\midrule
MMRL & 90.05 & 91.47 & 90.75 & 45.44 & 35.03 & 39.56 & 82.69 & 79.21 & 80.92 & 85.19 & 65.46 & 74.03 & 95.98 & 64.08 & 76.85 & 88.11 & 79.88 & 83.79 \\
\textbf{+PPOM} & 90.45 & 92.11 & \textbf{91.27} & 47.84 & 36.23 & \textbf{41.23} & 82.85 & 79.47 & \textbf{81.13} & 86.00 & 65.58 & \textbf{74.41} & 95.98 & 64.44 & \textbf{77.11} & 89.19 & 81.29 & \textbf{85.06} \\
\bottomrule
\end{tabular}

\end{table*}
 \begin{table*}[t]
    \caption{Cross-dataset transfer of PPOM across 4 hosts on ImageNet source and the other 10 evaluation-only target datasets.}
    \label{tab:cross}
    \centering
    \setlength\tabcolsep{5pt}
    \small

\begin{tabular}{c|c|ccccccccccc}
\toprule
\rowcolor{gray!10} {\cellcolor{gray!10}} & \textbf{Source} & \multicolumn{11}{c}{\textbf{Target}} \\
\rowcolor{gray!10} \multirow{-2}{*}{{\cellcolor{gray!10}}\textbf{Method}} & ImageNet & \textbf{Average} & Caltech & Pets & Cars & Flowers & Food & Aircraft & SUN & DTD & EuroSAT & UCF \\
\midrule
CoOp & 71.64 & 64.91 & \textbf{93.71} & 90.08 & 64.47 & \textbf{69.47} & 85.94 & 20.28 & 66.24 & 42.26 & 49.30 & 67.30 \\
{\cellcolor{green!20}}\textbf{+PPOM} & {\cellcolor{green!20}}\textbf{72.14} & {\cellcolor{green!20}}\textbf{65.32} & 93.67 & \textbf{90.24} & \textbf{65.43} & 69.22 & \textbf{86.51} & \textbf{21.00} & \textbf{66.67} & \textbf{42.38} & \textbf{49.77} & \textbf{68.36} \\
\midrule
KgCoOp & 70.86 & 65.36 & 93.96 & 90.35 & 65.99 & 70.89 & 86.41 & 23.13 & 67.17 & \textbf{47.99} & \textbf{40.47} & 67.20 \\
{\cellcolor{green!20}}\textbf{+PPOM} & {\cellcolor{green!20}}\textbf{71.53} & {\cellcolor{green!20}}\textbf{65.80} & \textbf{94.16} & \textbf{90.68} & \textbf{67.02} & \textbf{71.50} & \textbf{86.87} & \textbf{23.76} & \textbf{67.79} & 47.46 & 40.46 & \textbf{68.36} \\
\midrule
PromptSRC & 71.88 & 65.42 & 93.02 & \textbf{90.98} & 65.53 & \textbf{70.08} & 86.14 & 23.91 & 67.54 & 46.22 & 41.69 & 69.10 \\
{\cellcolor{green!20}}\textbf{+PPOM} & {\cellcolor{green!20}}\textbf{72.40} & {\cellcolor{green!20}}\textbf{66.05} & \textbf{93.55} & 90.84 & \textbf{66.38} & 70.04 & \textbf{86.80} & \textbf{25.47} & \textbf{68.08} & \textbf{46.93} & \textbf{42.68} & \textbf{69.76} \\
\midrule
MMRL & 73.58 & 66.72 & 94.20 & \textbf{91.14} & 66.21 & 72.88 & 85.34 & 26.19 & 67.36 & \textbf{45.57} & 49.85 & 68.46 \\
{\cellcolor{green!20}}\textbf{+PPOM} & {\cellcolor{green!20}}\textbf{74.28} & {\cellcolor{green!20}}\textbf{67.33} & \textbf{94.56} & 91.09 & \textbf{67.54} & \textbf{72.92} & \textbf{85.82} & \textbf{26.37} & \textbf{67.85} & 45.51 & \textbf{51.38} & \textbf{70.26} \\
\bottomrule
\end{tabular}

\end{table*}

\subsection{Architecture-derived Phase Orbit} \label{sec.2.2}

\noindent \textbf{Patch-phase Cell.} Let the ViT patch embedding have an even, non-overlapping kernel and stride $s$. Token membership depends on the image origin modulo $s$. Rather than enumerate all $s^2$ pixel phases, PPOM selects the representative midpoint $r=s/2$ on each axis. For ViT-B/16, $r=8$. $s$ is predefined by the ViT patch embedder.

\noindent \textbf{Antithetic Views.} Define the finite orbit
\begin{equation}
\mathcal O_s=\{(0,0),(0,\pm r),(\pm r,0),(\pm r,\pm r)\}.
\end{equation}
For sampled offset $\delta=(d_y,d_x)$, let $P_r(x)$ reflection-pad every boundary by $r$. With input height $H$ and width $W$, translation is the shape-preserving crop for obtaining offset views:
\begin{equation}
T_\delta(x)=P_r(x)[:,r+d_y:r+d_y+H,r+d_x:r+d_x+W].
\end{equation}
Reflection avoids the artificial constant bands introduced by zero-padding while keeping visual input compatible with the hosts' pre-processing path. The views are concatenated on the batch dimension and passed through the unchanged host, giving $\ell_\delta=\ell_H(T_\delta(x);p^*)$. The offsets depend on neither labels nor confidence, so the same operation applies to base, new, source, and target examples. Sampling the midpoint and its antithetic signs covers the two axial phase directions and their joint displacement with a fixed nine-view budget.

\subsection{Family-balanced Orbit Marginalization}  \label{sec.2.3}

\noindent \textbf{Directional Cancellation.} Opposite translations are paired into horizontal, vertical, and diagonal evidence families. To avoid over-weighting families with more sampled views (e.g., diagonal with 4 views), we marginalize evidence uniformly:
\begin{equation}
h=\tfrac12(\ell_{0,-r}+\ell_{0,r}),\quad
v=\tfrac12(\ell_{-r,0}+\ell_{r,0}),
\end{equation}
\begin{equation}
d=\tfrac14\!\sum_{d_y,d_x\in\{-r,r\}}\ell_{d_y,d_x}.
\end{equation}
Pairing treats the sign of a displacement as a nuisance and cancels directional first-order effects under a locally smooth logit response. It also prevents a single padded boundary from defining a phase-family prediction. This deterministic integration process can reduce sensitivity to patch-grid alignment.

\noindent \textbf{Family-balanced Margin.} With identity $i=\ell_{0,0}$, PPOM predicts on test sets of base or new tasks from
\begin{equation}
\bar\ell=\tfrac14(i+h+v+d),\qquad
\hat y=\arg\max_c\bar\ell_c.
\end{equation}
Equal family mass further avoids the $1{:}2{:}2{:}4$ bias of a flat nine-view mean. Logits are averaged before softmax, so each view retains the host's shared temperature. In addition, independently normalized probabilities would introduce view-specific calibration prior to integration. 

\noindent \textbf{Host-preserving Execution.} PPOM modifies only the test-time visual sampling and encoding of prompt learners. For a batch of size $b$, it processes all $9b$ views in one concatenated call, each through the host's native logit path, while leaving training parameters, optimizer, shots, and epochs unchanged. PPOM is therefore a \textit{\textbf{training-free, plug-and-play operator}} for prompt learning.

\section{Experiments}

\subsection{Experimental Setup}

\noindent \textbf{Datasets.} 
Following mainstream prompt learning approaches, we evaluate PPOM on ImageNet, Caltech101, Oxford Pets, Stanford Cars, Flowers102, Food101, FGVC Aircraft, SUN397, DTD, EuroSAT, and UCF101. PPOM only accesses the same
test sets available to each host, preventing data leakage.

\noindent \textbf{Host Models.} 
We integrate PPOM into four prompt learning hosts with distinct prompt designs and feature interaction mechanisms: CoOp~\cite{Zhou2022CoOp}, KgCoOp~\cite{Yao2023KgCoOp}, PromptSRC~\cite{Khattak2023PromptSRC}, and MMRL~\cite{Guo2025MMRL}, to evaluate its plug-and-play compatibility.

\noindent \textbf{Implementation Details.} 
All experiments apply CLIP ViT-B/16 as backbone and follow the 16-shot protocol. PPOM and tuned host model share the same checkpoints and parameters.

\subsection{Experimental Results}

\noindent \textbf{Base-to-New Generalization.}
Following the standard base-to-new protocol in prompt learners \cite{Li2025DPC, Yao2023KgCoOp, li2025mao}, the categories of each dataset are evenly divided into base and new subsets. All prompt-learning hosts are trained exclusively on the base-class training data, while the new classes remain unseen until evaluation. PPOM is then applied to the resulting checkpoint in evaluation-only mode, so its comparison with the native host is strictly checkpoint-paired and introduces no additional training data or optimization. We report Base accuracy, New accuracy, and their harmonic mean (HM). As shown in Table~\ref{tab:b2n}, PPOM improves the average HM of CoOp~\cite{Zhou2022CoOp}, KgCoOp~\cite{Yao2023KgCoOp}, PromptSRC~\cite{Khattak2023PromptSRC}, and MMRL~\cite{Guo2025MMRL} under the same training setup, demonstrating consistent host-level complementarity despite heterogeneous per-dataset changes.


\noindent \textbf{Cross-dataset Transfer.}
We fine-tune 4 host models on all ImageNet source classes and directly evaluate the same checkpoint on 10 target datasets under a zero-shot setting. The training-free PPOM is applied to both target and source datasets. As shown in Table~\ref{tab:cross}, across all hosts, PPOM improves both ImageNet source accuracy and average target performance, with gains on most targets. This indicates that its stride-derived phase integration transfers across datasets without target labels or dataset-specific coefficient selection. 



\subsection{Ablation Study}

\begin{table}[t]
    \centering
        \caption{Ablation study of components in PPOM+MMRL.}
        \label{tab:abl-module}
    \setlength{\tabcolsep}{6pt}
    \small
    \begin{tabular}{lcc|ccc}
        \toprule
        \rowcolor{gray!10} & \multicolumn{2}{c|}{\textbf{Modules}} & \multicolumn{3}{c}{\textbf{Average (11)}} \\
        \rowcolor{gray!10}  &Phase Orbit & Family Balance & Base & New & HM \\
        \midrule
        (1) & $\times$     & $\checkmark$        & 85.51 & 75.33 & 80.10 \\
        (2) & $\checkmark$ & $\times$        & 85.89 & 75.55 & 80.39 \\
        (3) & $\checkmark$ & $\checkmark$    & \textbf{86.18} & \textbf{75.91} & \textbf{80.72} \\
        \bottomrule
    \end{tabular}
\end{table}

\noindent \textbf{Validity of Proposed Components.}
Table~\ref{tab:abl-module} examines the contribution of each PPOM component. We discover that (1) removing the \textit{Phase Orbit} in Sec.~\ref{sec.2.2} by repeating the identity view nine times reduces PPOM to native MMRL performance, indicating that the gain of PPOM comes from exposing complementary patch phases. (2) Removing \textit{Family-balanced Orbit Marginalization} in Sec.~\ref{sec.2.3} by retaining all views but averaging them uniformly weakens the improvement of PPOM, confirming that balanced phase-family weighting is effective rather than view multiplicity alone. (3) The PPOM with full configuration performs best, which confirms the effectiveness of each component.

\begin{table}[t]
    \centering
        \caption{Ablation study of orbit family in PPOM+MMRL.}
        \label{tab:abl-family}
    \setlength{\tabcolsep}{6pt}
    \small
    \begin{tabular}{ccc|ccc}
        \toprule
        \rowcolor{gray!10} \multicolumn{3}{c|}{\textbf{Families}} & \multicolumn{3}{c}{\textbf{Average (11)}} \\
        \rowcolor{gray!10} Horizontal & Vertical & Diagonal & Base & New & HM \\
        \midrule
        $\times$     & $\checkmark$ & $\checkmark$ & 85.85 & 75.74 & 80.47 \\
        $\checkmark$ & $\times$     & $\checkmark$ & 85.63 & 75.59 & 80.30 \\
        $\checkmark$ & $\checkmark$ & $\times$     & 85.87 & 75.36 & 80.27 \\
        $\checkmark$ & $\checkmark$ & $\checkmark$ & \textbf{86.18} & \textbf{75.91} & \textbf{80.72} \\
        \bottomrule
    \end{tabular}
\end{table}



\noindent \textbf{Orbit Family.} 
In Table~\ref{tab:abl-family}, we remove each phase family in turn while retaining the identity and remaining families. Removing any family degrades overall HM, whereas the complete orbit achieves the strongest performance. This further indicates that different directions provide non-redundant evidence: axial and joint displacements induce distinct patch-token assignments, so combining all families captures complementary phase responses and enables more effective directional cancellation during marginalization.

\begin{figure}[t]
  \centering
  \includegraphics[width=\linewidth]{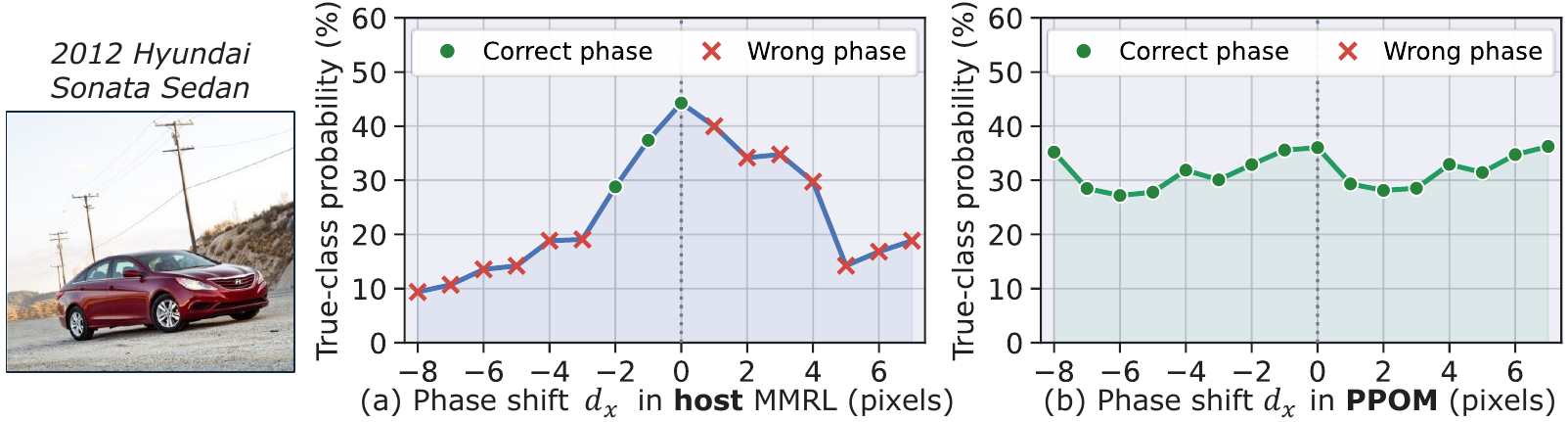}
  \caption{Case study of horizontal patch phase shift of ViT.}
  \label{Fig-casestudy}
\end{figure}

\noindent \textbf{Patch-Phase Robustness.} 
We further extend Fig.~\ref{Fig-eyecatch} with a case study in Fig.~\ref{Fig-casestudy}. Native MMRL prediction varies sharply across horizontal patch phases of ViT and frequently outputs incorrect predictions (\textit{13/16}), whereas PPOM maintains stable true-class confidence and correct predictions across shifts. This example further supports PPOM’s ability to suppress phase-dependent instability and improve overall robustness.



\begin{table}
    \caption{Computation cost of PPOM on ImageNet dataset.}
    \label{tab:compute-cost}
\centering
\setlength\tabcolsep{3.5pt}
\small
\begin{tabular}{c|cccc|c} 
\toprule
\rowcolor{gray!10} \textbf{Model}    & \begin{tabular}[c]{@{}c@{}}\textbf{Learnable}\\\textbf{Params}\end{tabular} & \begin{tabular}[c]{@{}c@{}}\textbf{Memory}\\\textbf{(MB)}\end{tabular} & \begin{tabular}[c]{@{}c@{}}\textbf{Base Task}\\\textbf{Time}\end{tabular} & \begin{tabular}[c]{@{}c@{}}\textbf{Effective}\\\textbf{FPS}\end{tabular} & \begin{tabular}[c]{@{}c@{}}\textbf{HM}\\\textbf{Acc.}\end{tabular}  \\ 
\midrule
CoOp           & 8K         & 8114.6 & 106m52s          & 12.48          & 72.27          \\
\rowcolor{green!20}
\textbf{+PPOM} & \textbf{+0} & \textbf{1609.2} & \textbf{+11m4s}   & \textbf{37.68} & \textbf{73.02} \\ 
\midrule
MMRL           & 4.99M      & 6913.8          & 105m26s          & 12.65          & 74.48          \\
\rowcolor{green!20}
\textbf{+PPOM} & \textbf{+0} & \textbf{3216.6} & \textbf{+17m58s}  & \textbf{23.15} & \textbf{75.20} \\ 

\bottomrule
\end{tabular}
\end{table}

\noindent \textbf{Computational Cost.} 
Table~\ref{tab:compute-cost} shows that PPOM adds no learnable parameters in the test-time stage and incurs substantially lower memory and task time costs than host fine-tuning, while improving HM. Although multi-view inference increases test-time computation, PPOM requires neither retraining nor test-time adaptation, keeping its overall computational overhead lower than adaptation-based methods.

\section{Conclusion}

Empirical analysis shows that prompt learning is sensitive to phase shifts in visual sampling. To reduce this sensitivity at test time, PPOM separates semantic prompt adaptation from the spatial phase of patch tokenization. Its stride-derived orbit and family-balanced marginal preserve the trained host while integrating complementary visual evidence. The consistent host performance enhancement, tuning-free target transfer, and plug-and-play characteristics together establish the patch phase as a useful post-training operator.

\section{Acknowledgements}
This work is supported by the China Scholarship Council (CSC) and the UTS Top-Up Scholarship. Liang Wang and Haoyang Li have equal 50\% contributions on this paper.

\bibliographystyle{IEEEbib}
\bibliography{references}

\end{document}